\documentclass{article}

\usepackage{arxiv}

\usepackage{xspace}
\usepackage{amssymb}
\usepackage[utf8]{inputenc} 
\usepackage[T1]{fontenc}    
\usepackage{hyperref}       
\usepackage{url}            
\usepackage{booktabs}       
\usepackage{amsfonts}       
\usepackage{nicefrac}       
\usepackage{microtype}      
\usepackage{lipsum}		
\usepackage{graphicx}
\usepackage{natbib}
\usepackage{bbm}
\usepackage{subfigure}
\usepackage{amsmath,bm}
\usepackage{algorithm,algpseudocode}
\usepackage[table]{xcolor}
\usepackage{doi}

\def\system{\textsc{GALC-SLR}\xspace}
\def\asl{ASL\xspace}
\title{Multi-Label Gold Asymmetric Loss Correction with Single-Label Regulators}

\author{{Cosmin Octavian Pene}\\
	Delft University of Technology\\
	Delft, The Netherlands\\
	\texttt{c.o.pene@student.tudelft.nl} \\
	\And
	{Amirmasoud Ghiassi} \\
	Delft University of Technology\\
	Delft, The Netherlands\\
	\texttt{s.ghiassi@tudelft.nl} \\
	\And
	{Taraneh Younesian} \\
	Delft University of Technology\\
	Delft, The Netherlands\\
	\texttt{t.younesian@tudelft.nl} \\
	\And
	{Robert Birke} \\
	ABB Research\\
    Baden-D\"attwil, Switzerland\\
	\texttt{robert.birke@ch.abb.com} \\
	\And
	{Lydia Y.Chen} \\
	Delft University of Technology\\
	Delft, The Netherlands\\
	\texttt{lydiaychen@ieee.org} \\
	
}

\date{}

\renewcommand{\headeright}{}

\newcommand{\pluseq}{\mathrel{+}=}
\newcommand{\diveq}{\mathrel{/}=}

\hypersetup{
pdftitle={A template for the arxiv style},
pdfsubject={q-bio.NC, q-bio.QM},
pdfauthor={David S.~Hippocampus, Elias D.~Striatum},
pdfkeywords={First keyword, Second keyword, More},
}

\begin{document}
\maketitle

\begin{abstract}
	Multi-label learning is an emerging extension of the multi-class classification where an image contains multiple labels. Not only acquiring a clean and fully labeled dataset in multi-label learning is extremely expensive, but also many of the actual labels are corrupted or missing due to the automated or non-expert annotation techniques. Noisy label data decrease the prediction performance drastically. In this paper, we propose a novel Gold Asymmetric Loss Correction with Single-Label Regulators (\system) that operates robust against noisy labels. \system estimates the noise confusion matrix using single-label samples, then constructs an asymmetric loss correction via estimated confusion matrix to avoid overfitting to the noisy labels. Empirical results show that our method outperforms the state-of-the-art original asymmetric loss multi-label classifier under all corruption levels, showing mean average precision improvement up to 28.67\% on a real world dataset of MS-COCO, yielding a better generalization of the unseen data and increased prediction performance.
\end{abstract}


\section{Introduction}
\label{sec:intro}
Real-world images naturally contain multiple objects corresponding to diverse labels, which elevates deep learning models for multi-label learning. Multi-label classification is an extension of multi-class classification where the input is not assigned only a single label, but multiple ones.

It is extremely time consuming and expensive to collect high quality labels for single-label images. Even long-standing and highly curated datasets, e.g. CIFAR~\citep{kriz-cifar10}, contain wrong labels~\citep{ChenLCZ19}. Multi-label learning, where each image has multiple possible label combinations, exacerbates this problem~\citep{evaluatingmllclassifiersnoisylabels}.
For instance, \citep{noisydatasetsminsup} shows that the Open Images dataset~\citep{openimages}, which is widely used for multi-label and multi-class image classification, contains 26.6\% false positives among the training label set.

\begin{figure}[t]
\centering
\hfill
\subfigure[Example of image with wrong labels]{
    \label{fig:wrong_labels}
    \includegraphics[width=0.52\linewidth]{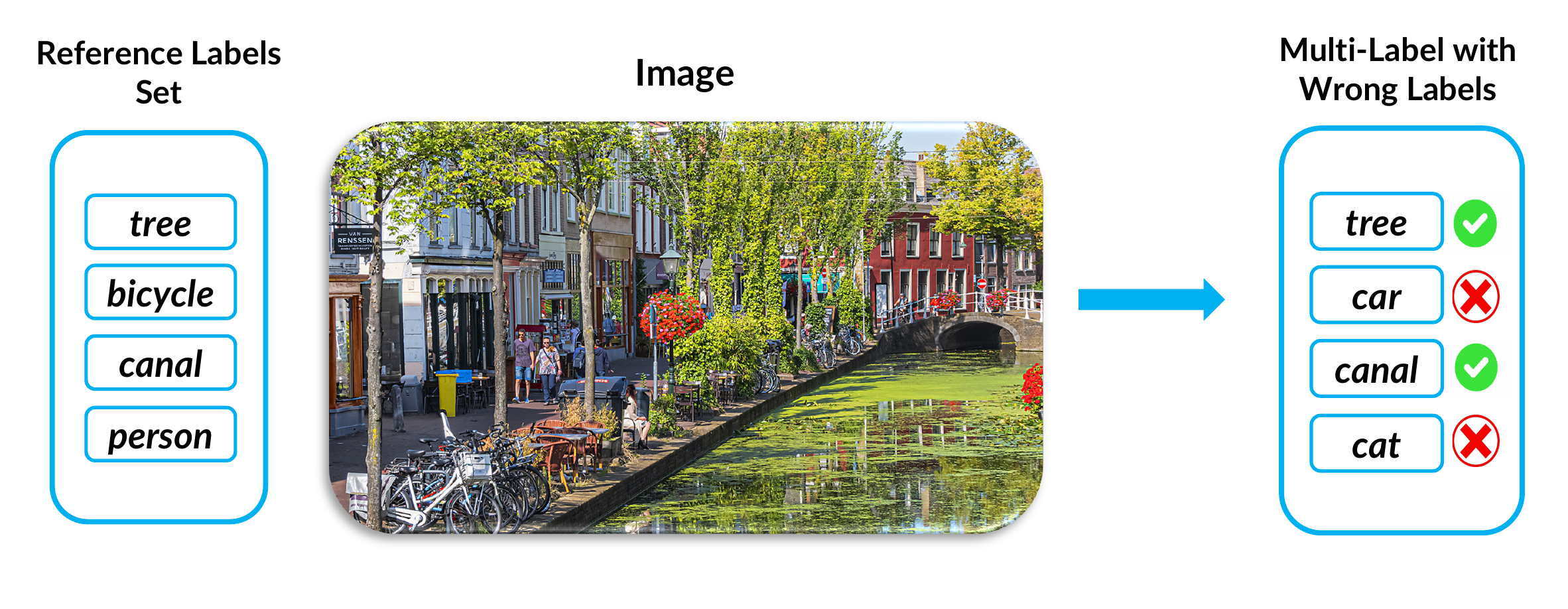}
}
\hfill
\subfigure[Impact of increasing wrong label ratios]{
    \includegraphics[width=0.38\linewidth]{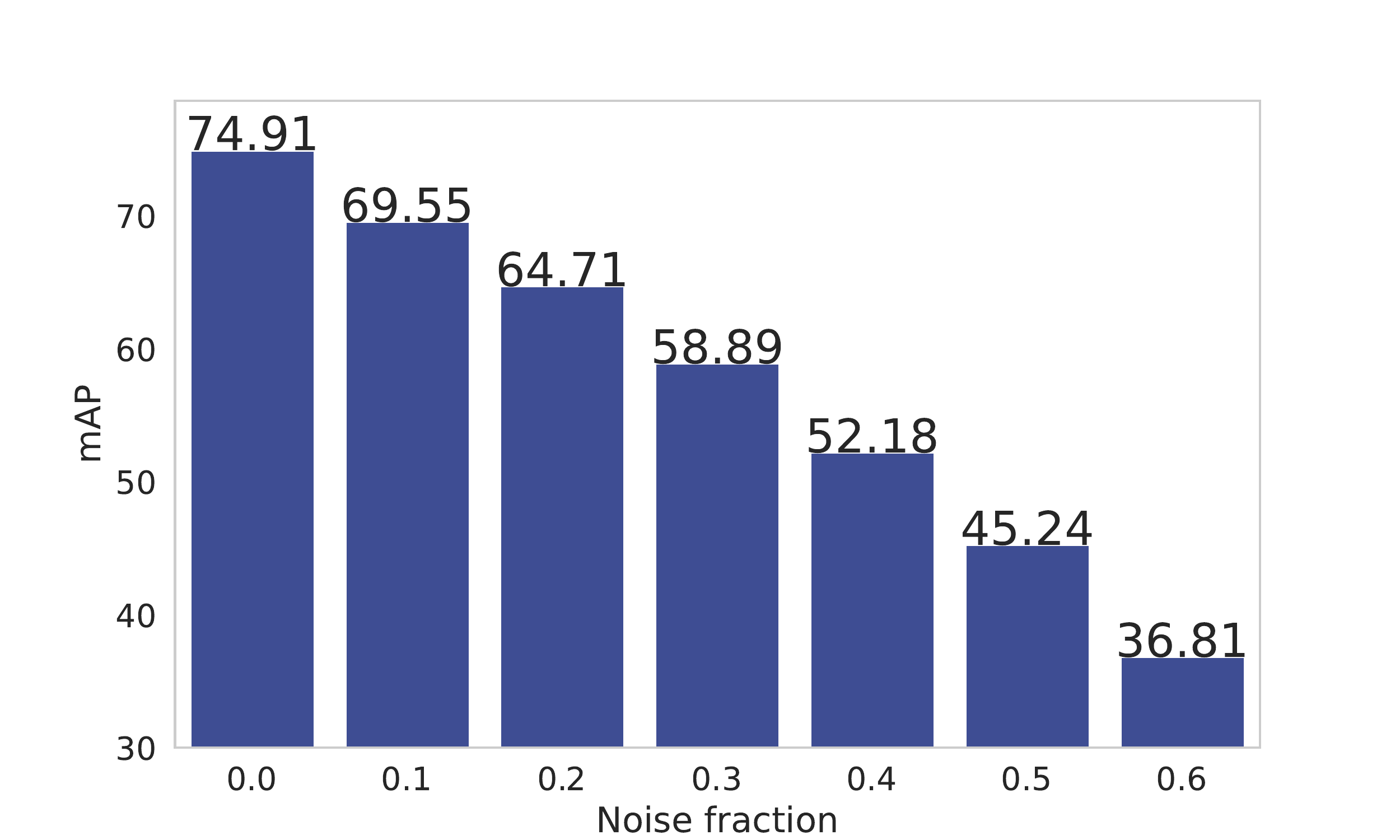}
    \label{fig:noise_impact_ASL}
}
\hfill
\caption{Wrong labels and their impact in multi-label classification.}
\label{fig:wrong}
\end{figure}

In single-label classification, label noise has been widely studied~\citep{surveyimgclassnoisylabels}. Due to the memorization effect Deep Neural Networks (DNNs) can overfit to noise degrading significantly their performance~\citep{ZhangBHRV17}. Several techniques have been proposed to counter the effect of wrong labels~\citep{glc,coteaching,losscorrectionapproach,cann}.
However, multi-label classification is a more complex problem. 
As depicted in Fig.~\ref{fig:wrong_labels}, each image comes with multiple labels including some wrong and some clean ones. This has a negative impact on the performance of DNNs.
According to~\citep{surveydnnnoisylabels}, existing noise resilient methods for single-label are not able to learn the correlation among multiple labels.
Even so little attention has been given to evaluating multi-label classifiers with noisy labels~\citep{evaluatingmllclassifiersnoisylabels}.

To fill the gap in noise-resilient multi-label classifiers, we propose Gold Asymmetric Loss Correction with Single-Label Regulators (\system). \system assumes that a small subset of the training samples can be trusted. We use this additional information to accurately estimate the noise corruption matrix. Due to class imbalance and label correlations, learning the noise in real-world multi-label datasets is more complicated than in real-world single-label datasets. Hence, we introduce a novel method that uses \textit{single label regulators} to rebalance the predictions towards a targeted label. This leads to accurate noise estimations used to correct the wrong labels during training  making the model robust to label noise even in the more challenging multi-label setting.


In comparison to the state-of-the-art Asymmetric Loss (ASL)  multi-label classifier~\citep{asymmetricloss} \system is significantly more accurate under label noise. ASL balances the probabilities of different samples by treating positive and negative samples differently, i.e. asymmetrically. In empirical evaluation on the MS-COCO dataset~\citep{mscoco} \system outperforms ASL under all tested noise ratios from 0\% to 60\%. \system improves the mean Average Precision (mAP) over ASL on average by 13.81\% and up to 28.67\%.

The contributions of this paper are summarized as follows:

\begin{itemize}
    \item We design a noise estimation technique that uses trusted multi-label and single-label data in order to calculate the corruption matrix.
    \item Using our noise estimation, we design a robust multi-label classifier, \system, based on a loss correction approach.
    \item We compare \system against a state-of-the-art classifier under noisy labels and study the behaviour of \system in target ablation study experiments.
\end{itemize}


\subsection{Motivation example}

Our motivation stems from the detrimental effects label noise in training data can have on the model performance. We demonstrate this using the state-of-the-art ASL~\citep{asymmetricloss} method to train a TResNet-M~\citep{Ridnik2021WACV} network on the MS-COCO dataset~\citep{mscoco}. ASL applied on TResNet ranks top on the leader board for MLC on MS-COCO\footnote{\url{https://paperswithcode.com/sota/multi-label-classification-on-ms-coco} visited June 24, 2021.}.
We inject symmetric label noise (details in Section~\ref{sec:experiment_setup}) at various corruption levels, from 0\% to 60\%, and report the mean average precision to asses the impact of wrong labels. mAP is considered by many recent works~\citep{genmltransf,mlgcn} as important metric for performance evaluation in multi-label classification, since it takes into consideration both false-negative and false-positive rates~\citep{asymmetricloss}.
Fig.~\ref{fig:noise_impact_ASL} shows the results: each additional 10\% noisy labels leads to a 5\%-8\% reduction in mAP score. 
Since it is hard and costly to avoid label noise~\citep{evaluatingmllclassifiersnoisylabels}, it is vital to develop robust classifiers that can avoid overfitting to the label noise in the training data.

\section{Related Work}
\label{secrelated}
Recent literature has shown increased interest towards robustness against noisy labels in training, much more in single-label classification than in multi-label learning.
We first investigate the robust learning solutions in single-label classification, followed by an analysis of the multi-label context.

{\bf Single-label Classification}
Many researchers have been tackling the problem of noisy labels in single-label classification.
\citep{surveydnnnoisylabels} distinguishes between five categories of robust DNNs. Some classifiers such as C-model~\citep{cmodel}, Contrastive-Additive Noise Network~\citep{cann} and Robust Generative Classifier~(RoG)~\citep{rog} rely on a noise adaptation layer at the top of the softmax layer. Another solution is to use regularization techniques such as data augmentation~\citep{dataaug}, weight decay~\citep{weighdecay}, dropout~\citep{dropout} and batch normalization~\citep{batchnorm}. These methods perform well on low to moderate noise, but fail on datasets with higher noise~\citep{regulestimation}. Other works use Sample Selection showing impressive results even under heavy noise. Examples are MentorNet~\citep{mentornet}, Co-teaching~\citep{coteaching} and Co-teaching+~\citep{coteaching+}. Another stream of work estimates the noise corruption matrix to correct the labels during training, without, e.g. Forward~\citep{losscorrectionapproach} and Masking~\citep{masking}, or with, e.g. Golden Loss Correction~\citep{glc}, a small portion of trusted, i.e. verified clean, data. 

{\bf Multi-label Classification}
In multi-label learning, little attention has been given to the consequences of label noise~\citep{surveydnnnoisylabels}. Few papers treat noisy labels in the multi-label context.
\citep{SunFWLJ19} learns by a low-rank and sparse decomposition approach to obtain ground-truth and irrelevant label matrices.
\citep{pmll} introduces label confidence to recover the true labels. 
\citep{XieH20} formalizes the two objective of recovering ground-truth and identify noisy labels via a unified regulators-based framework.
The topic is gaining traction with some more works available as technical reports.
\citep{evaluatingmllclassifiersnoisylabels} leverages context to identify noisy labels.
\citep{li2020} proposes a two-step noise correction.
\citep{mpvae} acknowledges and evaluates the impact of noisy labels even if it is not the main goal of the authors.

In contrast to the above, \system aims to leverage single-label regulators together with a small fraction of trusted data to avoid overfitting to noisy labels in  multi-label classification.

\section{Methodology}
\subsection{Notation}

Consider the multi label dataset $\mathcal{D} = \{(\boldsymbol{x}_i, \tilde{\boldsymbol{y}}_i)\}_{i = 1}^N$  where $\boldsymbol{x}_i \subset \mathbb{R}^d$ denotes the $i^{th}$ sample out of $N$ with $d$ features. $\tilde{\boldsymbol{y}}_i \subset [0, 1]^K$ denotes the corresponding label vector 
over $K$ classes. The label vector is affect by noise, hence $\tilde{\boldsymbol{y}}$ can be clean ($\boldsymbol{y}$) or noisy ($\hat{\boldsymbol{y}}$).
Similar to GLC~\citep{glc}, we assume that a subset of the data, i.e. \textit{gold dataset} $\mathcal{G} \subset \mathcal{D}$, can be trusted. $|\mathcal{G}|$ contains samples $(\boldsymbol{x}, \boldsymbol{y})$ with no corrupted labels. We refer to the rest of the samples $(\boldsymbol{x}, \hat{\boldsymbol{y}})$ with potentially corrupted labels as \textit{silver dataset} $\mathcal{S} = \mathcal{D} - \mathcal{G}$. We define the \textit{trusted fraction} as the ratio $\frac{|\mathcal{G}|}{|\mathcal{G}| + |\mathcal{S}|}$. Furthermore, we assume that a small dataset of clean single-label images $\mathcal{G}_S$ is available. 
We use these sets to train a \emph{silver} $f(.;\theta)$ and a \emph{gold} $g(.;\phi)$ classifier and estimate the noise given by a $K \times K$ noise corruption matrix \textbf{C}. The elements $\boldsymbol{C}_{ij}$ are the probability of label $i$ to be flipped into label $j$, formally:
\[
\boldsymbol{C}_{ij} = p(\widehat y_{j}=1 \wedge \widehat y_{i}=0 | y_{j}=0 \wedge y_{i}=1)
\]

\subsection{Overview of GALC-SLR}

We propose a novel \textit{Gold Asymmetric Loss Correction with Single-Label Regulators} training approach. \system combines an asymmetric loss approach with a gold loss correction approach to counter noisy labels. The asymmetric loss treats relevant and irrelevant labels differently and has been shown to obtain impressive results on several MLC datasets~\citep{asymmetricloss}.  
The gold loss correction (GLC) assumes that a small subset of trusted samples is available to accurately estimate the true corruption matrix. It is a powerful method that achieves impressive results under both symmetric and asymmetric noise in the single-label setting~\citep{glc}. 
The original GLC assumes conditional independence of $\boldsymbol{y}$ given $\boldsymbol{x}$. This assumption holds when $\boldsymbol{y}$ is deterministic in $\boldsymbol{x}$. This does  not hold for multi-label classification, because an image can have multiple labels. Furthermore, in multi-label classification there can be label correlations that the original formula does not take into account, making it impossible to target a specific label. 
To derive an accurate multi-label, rather than single-label, noise corruption matrix \system uses single-label regulators and sigmoid classification which gives a more reliable representation of the noise in the multi-label context.

\begin{figure}[t!]
\centering
\includegraphics[trim=0 30 0  0,clip,width=0.8\columnwidth]{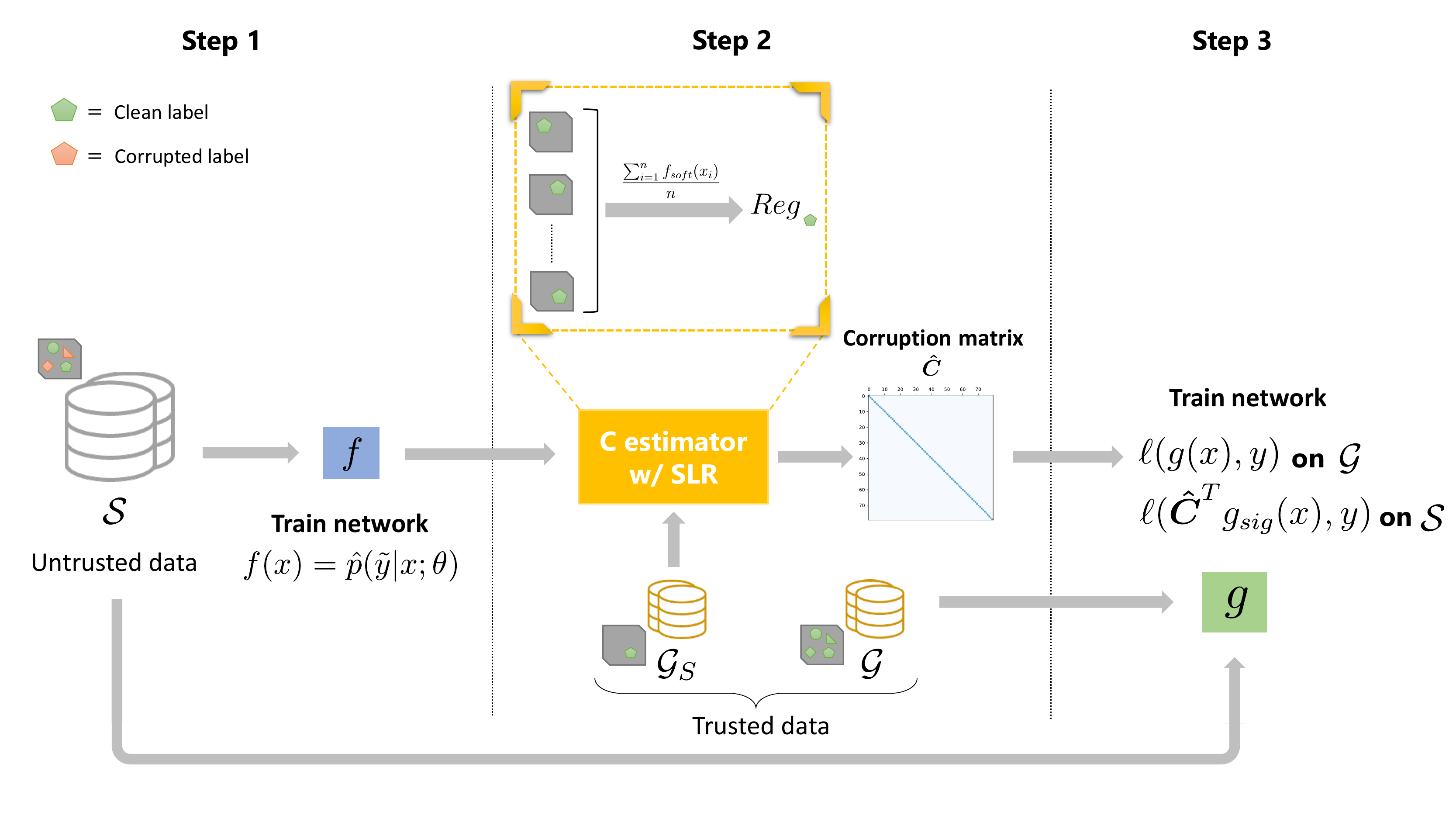}
\caption{Overview of GALC-SLR}
\label{fig:arch_galc}
\end{figure}

Figure~\ref{fig:arch_galc} presents an overview of the \system method. It includes three steps. Step~1: we train a classifier $f(.;\theta)$ using ASL loss on the noisy samples in $\mathcal{S}$. Step~2 is the heart of \system. We use $f$ with the trusted samples in $\mathcal{G}$ to estimate the noise corruption matrix and correct it via single-label regulators derived via the samples in $\mathcal{G}_S$. Step~3: we train the final classifier $g(.;\phi)$ using ASL loss on samples from $\mathcal{G}$ and corrected samples from $\mathcal{S}$.

\subsection{Noise Corruption Matrix Estimation}

First, we train a \textit{silver classifier} $f(\boldsymbol{x};\theta) = \widehat p(\hat{\boldsymbol{y}} | \boldsymbol{x})$ on $\mathcal{S}$, using the asymmetric loss from ~\citep{asymmetricloss}:
\[
    \mathcal{L}_{ASL} = \left\{
  \begin{array}{@{}ll@{}}
    L_{+} = (1 - p)^{\gamma_{+}} \log(p)\\
    L_{-} = (p_m)^{\gamma_{-}}\log(1 - p_m)
  \end{array}\right.
\]
where $L_+$ and $L_-$ are the positive and negative loss parts used for relevant and irrelevant labels, respectively. $p$ is the network output probability and $\gamma_{+}, \gamma_{-}$ are the focusing parameters. Finally, $p_m = \max (p - m, 0)$
denotes the shifted probability by a margin hyper parameter $m$. 
Given the labels in $\mathcal{S}$ are potentially corrupted, $f$ is not a reliable classifier for our final predictions. However, we can use $f$ to estimate our multi-label corruption matrix $\widehat{\boldsymbol{C}}$.

\begin{algorithm}[t]
\caption{GALC-SLR $\mathbf{\widehat C}$ estimation}
\label{alg:galc_c_hat_estimation}
\begin{algorithmic}[1]
\State \textbf{Input:} \text{Untrusted data $\mathcal{S}$, trusted single-label data $\mathcal{G}_{S}$, silver classifier $f$}
\State \textbf{Output:} \text{Estimated  $\widehat{\boldsymbol{C}}$}
\State \text{Fill} $Reg \in \mathbb{R} ^{K \times K}$ \text{with zeros} \text{\textcolor{lightgray}{\textit{/* Calculate Regulator for each label */}}}
\For{$k = 1,...,K$} 
    \State $num\_examples = 0$
    \For{$(\boldsymbol{x}_{i}, \boldsymbol{y}_{i}) \in \mathcal{G}_{S}$ \text{such that} $\boldsymbol{y}_{ik} = 1$} \text{\textcolor{lightgray}{/* \textit{Mean of silver \textbf{softmax} predictions} */}}
        \State $num\_examples \pluseq 1$
        \State $Reg_{k\bullet} \pluseq f_{soft}(\boldsymbol{x}_{i})$
    \EndFor
    \State $Reg_{k\bullet} \diveq num\_examples$
\EndFor
\State \text{Fill} $\widehat{\boldsymbol{C}} \in \mathbb{R} ^{K \times K}$ \text{with zeros} \text{\textcolor{lightgray}{\textit{/* Estimate multi-label corruption matrix */}}}
\For{$k = 1,...,K$} 
    \State $num\_examples = 0$
    \For{$(\boldsymbol{x}_{i}, \boldsymbol{y}_{i}) \in \mathcal{S}$ \text{such that} $\boldsymbol{y}_{ik} = 1$} \text{\textcolor{lightgray}{\textit{/* For each image that contains label k */}}}
        \State $num\_examples \pluseq 1$, $num\_other\_labels = 0$
        \State \text{Fill} $regulators \in \mathbb{R} ^{K}$ \text{with zeros}
        \For{$p = 1,...,K$ \text{such that} $p\neq k \And \boldsymbol{y}_{ip} = 1$} \text{\textcolor{lightgray}{\textit{/* Sum up other labels */}}}
            \State $num\_other\_labels \pluseq 1$
            \State $regulators \pluseq Reg_{p\bullet}$
        \EndFor
        \State $\widehat{\boldsymbol{C}}_{k\bullet} \pluseq f_{sig}(\boldsymbol{x}_{i}) - regulators$ \text{\textcolor{lightgray}{\textit{/* Correct sigmoid prediction via regulators */}}}
        \State $\widehat{\boldsymbol{C}}_{k\bullet} \pluseq Reg_{k\bullet} * num\_other\_labels$ \text{\textcolor{lightgray}{\textit{/* Rebalance towards target label $k$*/}}}
    \EndFor
    \State $\widehat{\boldsymbol{C}}_{k\bullet} \diveq num\_examples$
\EndFor
\State $\widehat{\boldsymbol{C}} = sig(\widehat{\boldsymbol{C}})$ \text{\textcolor{lightgray}{\textit{/* Final scaling */}}}
\end{algorithmic}
\end{algorithm}

Algorithm~\ref{alg:galc_c_hat_estimation} depicts our novel multi-label corruption matrix estimation. First, we calculate the \textit{single-label regulators} by taking the average of our silver softmax predictions for each label (lines 3-10):
\[
    Reg_{k\bullet} = \frac{\sum_{i=1}^N f_{soft}(\boldsymbol{x_i})}{N}, \,\,\, \forall k \in K
\]
where $Reg \subset \mathbb{R}^{K\times K}$ and $Reg_{k\bullet}$ denotes the $k^{th}$ matrix row.
This is the main step in our method. It not only allows to target a specific label $k$ for its noise corruption estimation, but also to regulate the label correlations from the multi-label images. The next step is to explicitly estimate the noise corruption matrix. For each label $k$, we sum the rows in $Reg$ for the labels which are present in the image except $k$ (line 18-21).
Next, for the each row of the noise corruption matrix, we regulate the sigmoid predictions of $f$ by subtracting the summed regulators (line 22). Finally, we rebalance each row using the number of other labels in the image (line 23) and take the average over the number of samples (line 25) scaled via a sigmoid (line 27).   
Fig.~\ref{fig:corruption_matrix} compares the multi-label corruption matrix estimated by \system for 40\% symmetric label noise against the injected --ground truth-- one, and the GLC estimated one. We observe that \system's estimation is more resistant to imbalanced data with respect to GLC's estimation. This can be seen from the darker, closer to the truth, diagonal values and the more pronounced difference with respect to the off-diagonal values. Note that to highlight this effect we avoid the last sigmoid scaling for better contrast in the figures.


With the estimated noise corruption matrix $\mathbf{\widehat{C}}$, we finally train the robust \textit{gold} classifier $g(.;\phi)$. We correct labels of the samples in $\mathcal{S}$ via $\mathbf{\widehat{C}}$ while leveraging samples in $\mathcal{G}$ as is. The loss function follows as:
\[
\begin{array}{@{}lll@{}}
 \ell &= \mathcal{L}_{ASL}(\widehat{\boldsymbol{C}}^{T}g_{sig}(\boldsymbol{x}), \hat{\boldsymbol{y}}),\,\,\, &\forall x \in \mathcal{S}\\
 \ell &= \mathcal{L}_{ASL}(g(\boldsymbol{x}),\boldsymbol{y})\,\,\, &\forall x \in \mathcal{G}.
\end{array}
\]




\label{sec:method}

\section{Evaluation}
\label{sec:eval}
\subsection{Experiment setup}
\label{sec:experiment_setup}

{\bf Datasets}.
We evaluate \system using  MS-COCO~\citep{mscoco} dataset. 
MS-COCO is a popular real-world dataset of common objects in context widely used for evaluation of multi-label classification. The training and validation datasets contain 82K and 40K images, respectively. Each image is tagged on average with 2.9 labels belonging to 80 classes.

To explicitly test \system in the more challenging multi-label setting, we remove all single label images from both the training and validation datasets. This does not only elicit a more reliable evaluation, but it also allows to collect single-label samples to construct $\mathcal{G}_{S}$. The number of images per class label varies from 1, for unpopular classes, to 1234, for the most popular class with an average of 210.2 images per class. 
More single-label samples allow to estimate more accurate regulators which in turn leads to a more robust classifier. 
Finally, we split the training data into a \textit{gold} ($\mathcal{G}$) and a \textit{silver} ($\mathcal{S}$) datasets. As base we use 10\% as gold data, leading to 6.5K clean samples and 58.7K samples injected with noisy labels.

{\bf Label Noise}.
Label noise in multi-label data is more complex than in a single-label context since each sample has an arbitrary number of labels. We follow previous works~\citep{mentornet,losscorrectionapproach} and inject symmetric noise, but with an extra step. Specifically, we select a fraction $\eta$, i.e. the noise ratio, of labels and flip them to another class with uniform probability. This corresponds to a noise corruption matrix having elements $C_{ij}$ as follows:
\[
C_{ij} = 
  \Bigg\{ \begin{array}{rl}
    1 - \eta & \text{if}\; i = j\\
    \dfrac{\eta}{K - 1} & \text{if}\; i \neq j \\
  \end{array}
\]
In order to ensure wrong label injection, we test whether or not the new label is already associated with the image. If it does, we repeatedly elect a new label until we select one which is not yet present. In order to evaluate how robust \system is to noise, we test our method against multiple noise ratios --from 0\% to 60\%.
\begin{figure}[t]
\centering
{
	\hfill
	\subfigure[True]{
	    \label{subfig:true_corruption_matrix}
	    \includegraphics[trim=80 30 90 20,clip,width=0.23\textwidth]{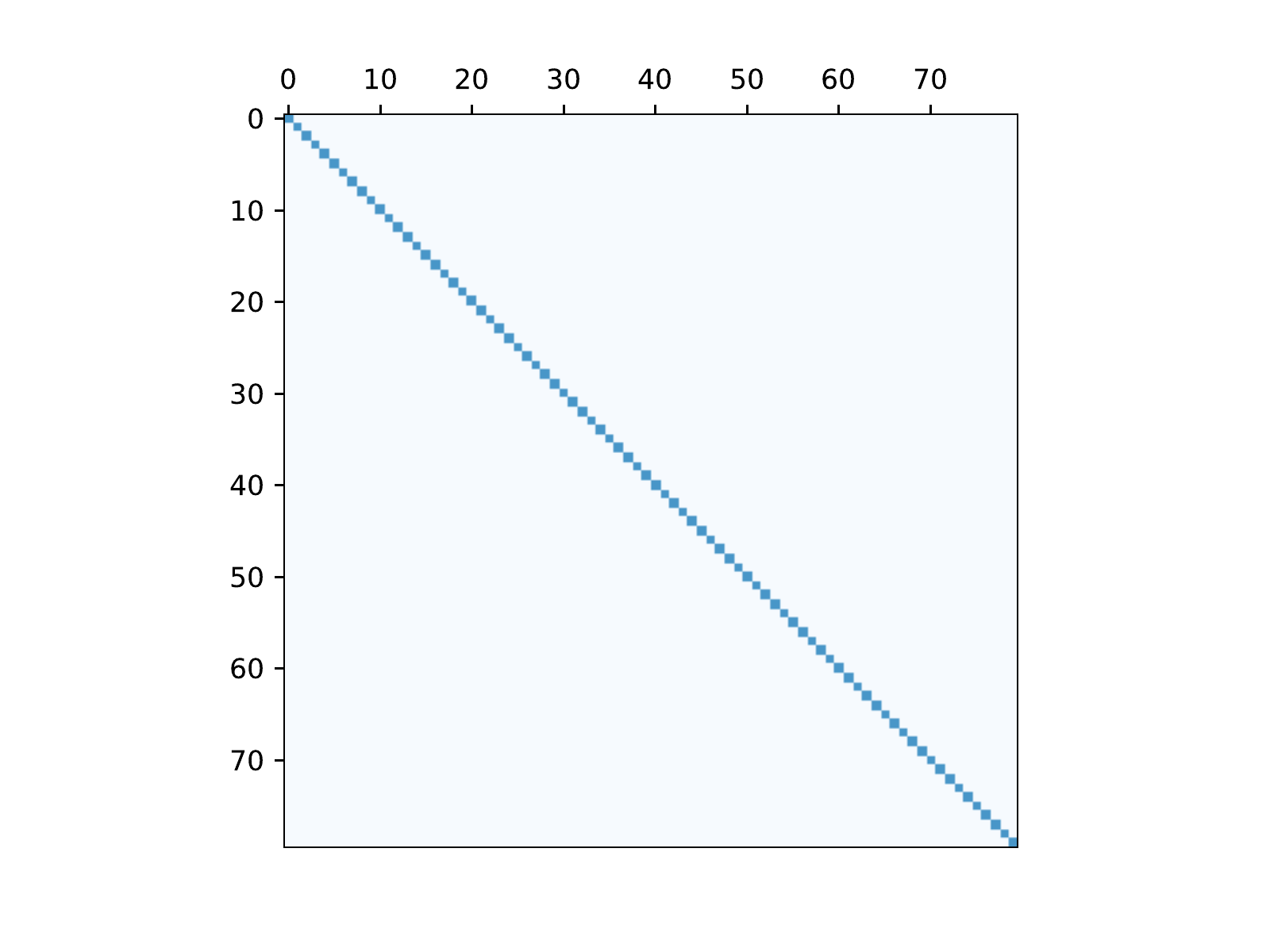}}
	\hfill
	\subfigure[GALC-SLR]{
	    \label{subfig:estimate_corruption_matrix}
	    \includegraphics[trim=80 30 90 20,clip,width=0.23\textwidth]{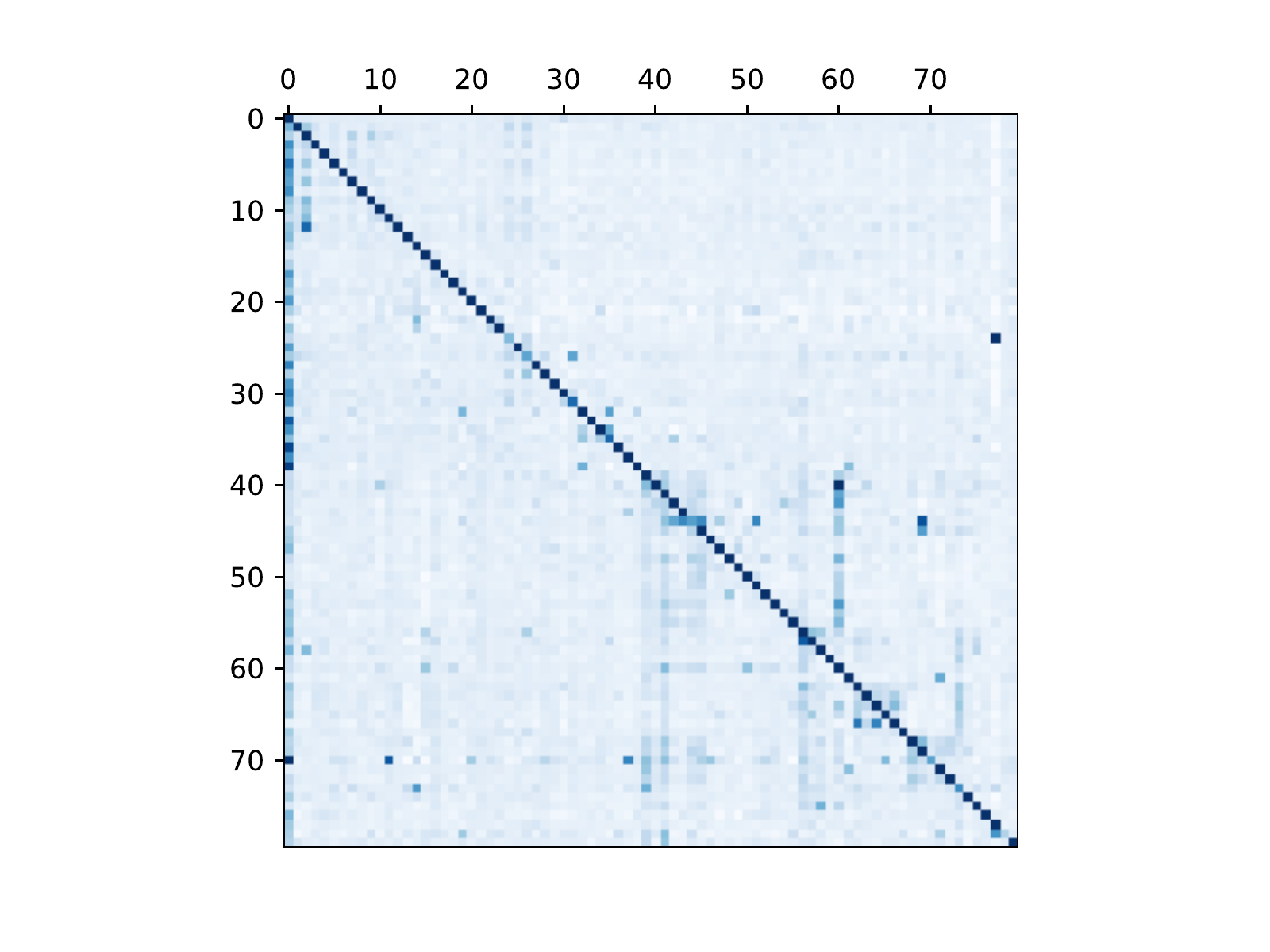} }
	\hfill
	\subfigure[GLC]{
	    \label{subfig:estimate_corruption_matrix_glc}
	    \includegraphics[trim=80 30 90 20,clip,width=0.23\textwidth]{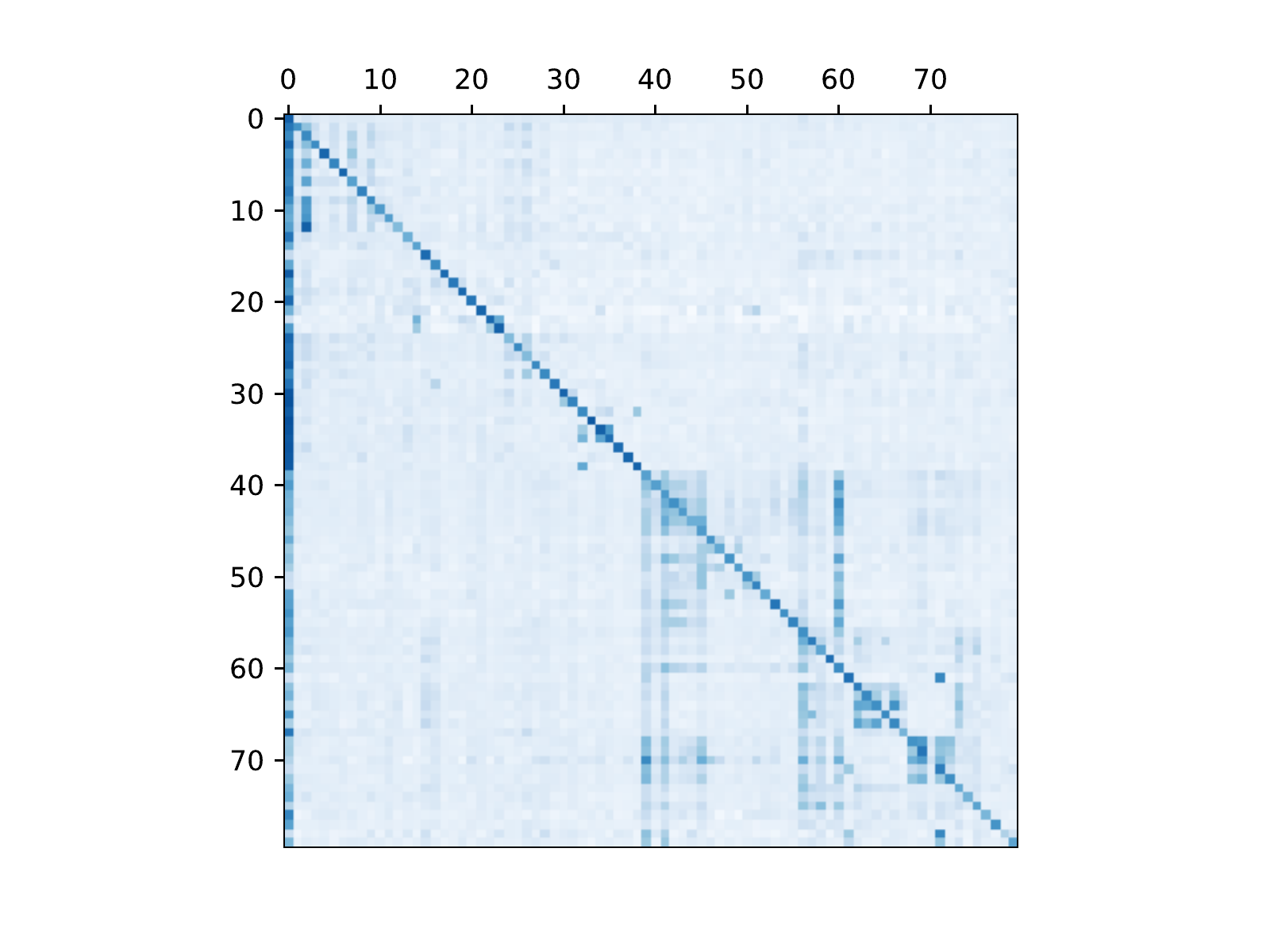} }
	\hfill
}
\caption{Comparison between multi-label corruption matrices with 40\% noise.}
\label{fig:corruption_matrix}
\end{figure}

{\bf Evaluation Metrics}.
For a comprehensive and reliable evaluation, we follow conventional settings and report the following metrics: mean average precision (mAP), average per-class F1 (CF1) and average overall F1 (OF1). These metrics have been widely used in literature to evaluate multi-label classification~\citep{asymmetricloss,surveydnnnoisylabels,mapimportance}. and have been shown to dramatically decrease with label noise~\citep{evaluatingmllclassifiersnoisylabels}. Note that only the training set is affected by noise, whereas the evaluation metrics are computed on the clean testing set.

{\bf DNN Architecture}.
As base architecture for the DNN we use TResNet~\citep{Ridnik2021WACV}. 
TResNet network is a high performance GPU-dedicated architecture based on ResNet50 designed to increase the model prediction performance, without increasing training or inference time. The TResNet network is pre-trained on the ImageNet-21K dataset for better generalizability and increased prediction accuracy~\citep{imagenet21k}. In particular we use the TResNet-M version.

{\bf Baseline}.
As baseline we compare against ASL~\citep{asymmetricloss} using the code provided by the authors.
\system assumes access to a small subset of clean samples.
For a fair comparison, we test ASL and \system on the same datasets with the same label noise. The only additional knowledge of our method is which labels are trusted, i.e. belonging to the small golden dataset $\mathcal{G}$, and which are potentially corrupted.

\textbf{Implementation Details}.
We use PyTorch v1.9.0 for both \system and ASL, and the default parameters provided in~\citep{asymmetricloss} except that we always take the last trained model due to the \textit{memorization effect}. The number of training epochs is an important parameter for a reliable evaluation, especially in a noisy setting. DNNs are shown to present the so-called \textit{memorization effect}~\citep{memorization1,memorization3,memorization4} benefiting in general from this factor to achieve a better prediction performance in atypical samples. However, \citep{memorization2} suggests that with noisy data, DNNs prioritize learning simple patterns first. From preliminary experiments we see that 80 epochs are enough for the learning to stabilize.


\subsection{Results}
In this subsection we empirically compare the performance of \system to the performance of ASL under 0\% to 60\% symmetric noise. We aim to show the effectiveness of our \system in robustly learning from noisy data.

\begin{figure}[t]
\centering
\subfigure[Mean Average Precision (mAP)]{
     \label{subfig:barplot_map}
     \includegraphics[trim=35 0 70 20,clip,width=0.23\columnwidth]{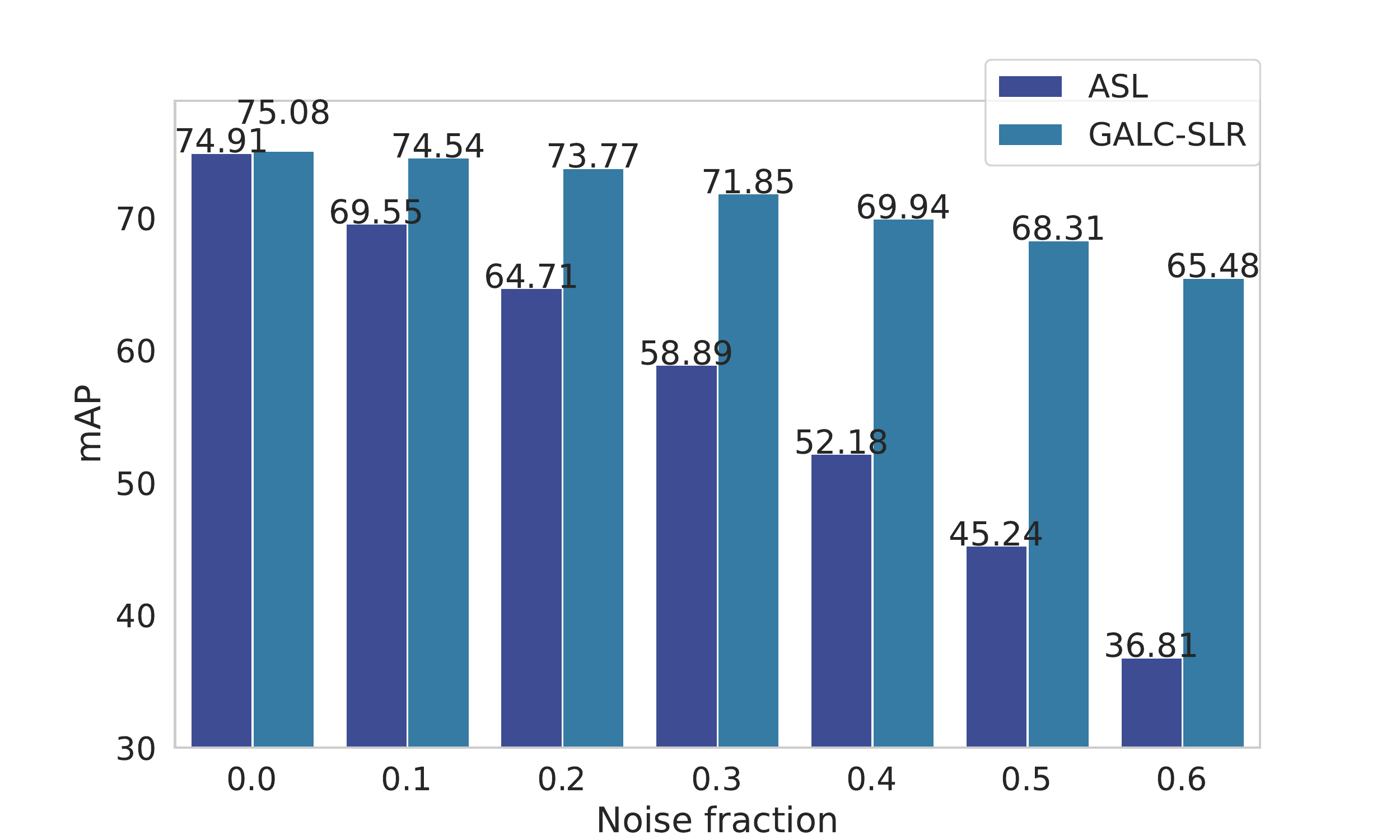}
     }
\hfill
\subfigure[Average per-class F1 (CF1)]{
     \label{subfig:barplot_cf1}
     \includegraphics[trim=35 0 70 20,clip,width=0.23\columnwidth]{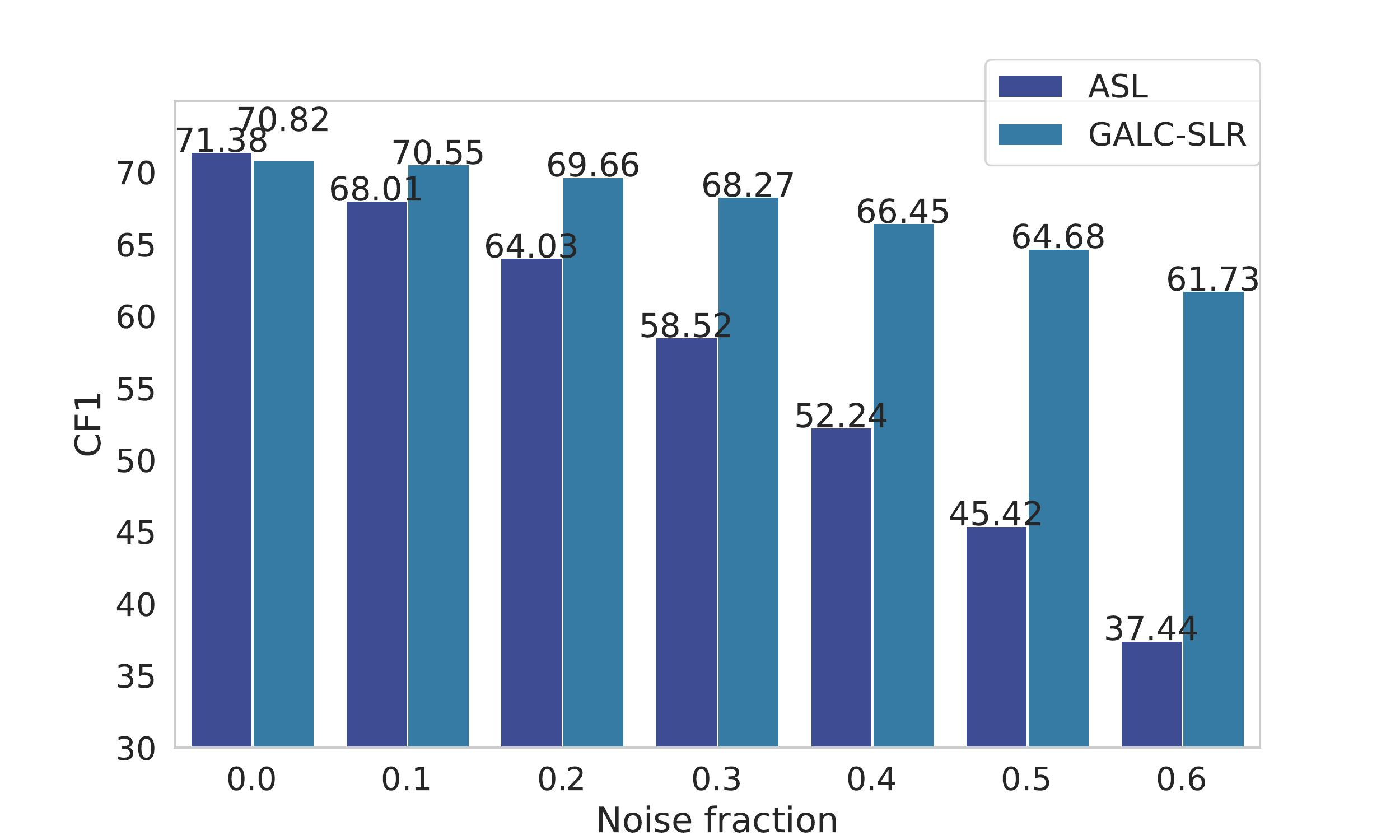}
     }
\hfill
\subfigure[Average overall F1 (OF1)]{
     \label{subfig:barplot_of1}
     \includegraphics[trim=35 0 70 20,clip,width=0.23\columnwidth]{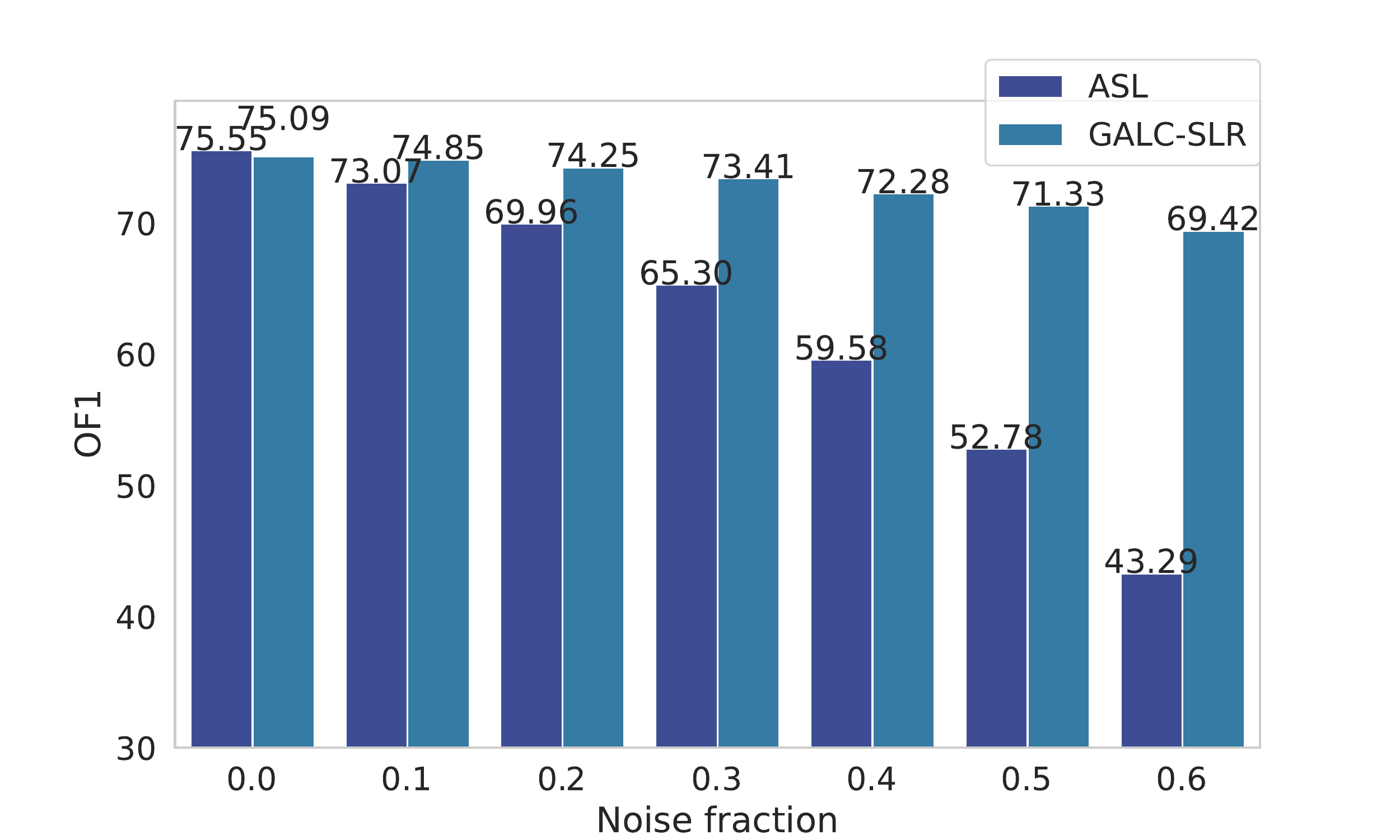}
     }
\hfill
\subfigure[Memorization effect under 40\% noise]{
     \label{subfig:memorization_effect_all_no_true}
     \includegraphics[width=0.23\columnwidth]{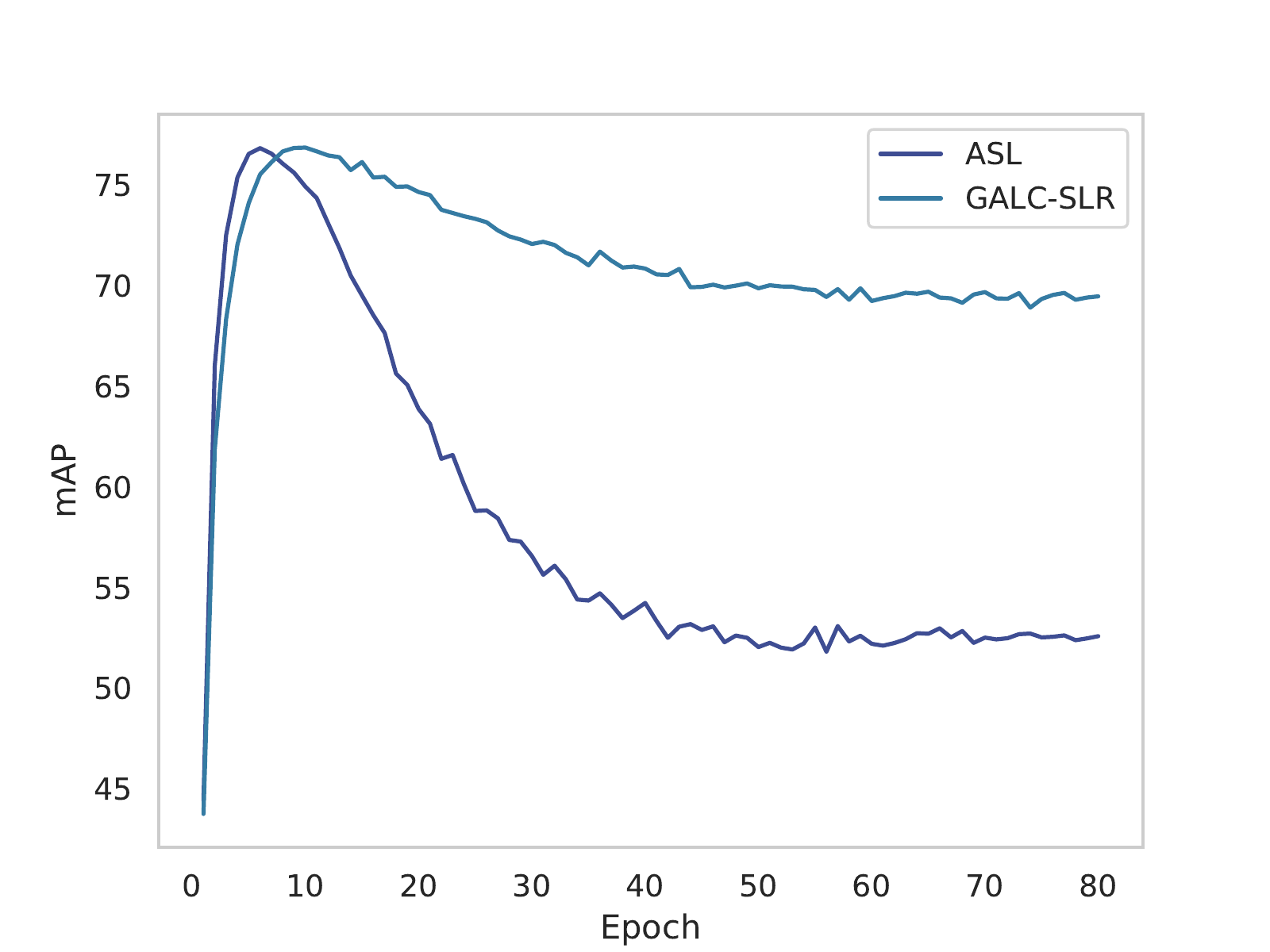}
     }     
\caption{Evaluation of \system and \asl on MS-COCO with symmetric label noise.}
\label{fig:results}
\end{figure}

Fig.~\ref{fig:results} shows the comparison results.
The performance of both systems decreases under increasing noise levels, but \system is significantly more robust.
In terms of mAP \system consistently outperforms ASL for all noise ratios (see Fig.~\ref{subfig:barplot_map}). 
ASL's performance drops an average of 5.34\% points with each 10\% noise, while \system's performance decreases with only 1.07\% points. Under severe noise, i.e. 60\%, the gap between \system and ASL is more than 28\% points and only 9.6\% points worse than without noise. In comparison ASL drops by 38.1\% points from 0\% to 60\% noise. This shows that \system is robust even to high noise levels. Similar results apply for both CF1 and OF1, see Fig.~\ref{subfig:barplot_cf1} and Fig.~\ref{subfig:barplot_of1}, respectively. Even if ASL is slightly better in the no noise case, the performance quickly degrades with additional noise. At 60\% \system is better by 24.3\% and 26.1\% points for CF1 and OF1, respectively. 


To reliably assess the correctness of \system, we also investigate the observed memorization effect for \system (depicted in Fig.~\ref{subfig:memorization_effect_all_no_true}). Both \system and \asl follow the same trend. First they learn the easy patterns, achieving a high accuracy after just a few epochs. However, afterwards the performance slowly degrades over training effort and finally stabilizes after 60 epochs.
The figure clearly shows the advantage of \system over \asl in the different levels at which they plateau. Moreover, one can observe that \system has a slight delay in learning at the beginning of the training, i.e. \system peaks at epoch 10, while \asl at epoch 6.
This observation indicates that \system does not help in terms of learning speed nor in reaching a higher performance during training, but by preventing to overfit to the noisy labels. This makes the DNN more resistant to wrong label information. This suggests that our method can also be applied to other existing classifiers and domains.


\subsection{Ablation Study}
To better understand the performance of \system, we perform extra ablation studies to investigate the effects of: i) errors in the noise corruption matrix estimation; ii) impact of the gold dataset size (both studied in experiment I); and iii) impact of number of single-label images (studied in experiment II). The base setup of the experiments is the same as in Section~\ref{sec:experiment_setup} with the only changes specifically mentioned. 

\begin{figure}[ht]
\centering
{
	\hfill
	\subfigure[Impact of trusted data]{
	    \label{subfig:ablation_study_combined}
	    \includegraphics[width=0.45\textwidth]{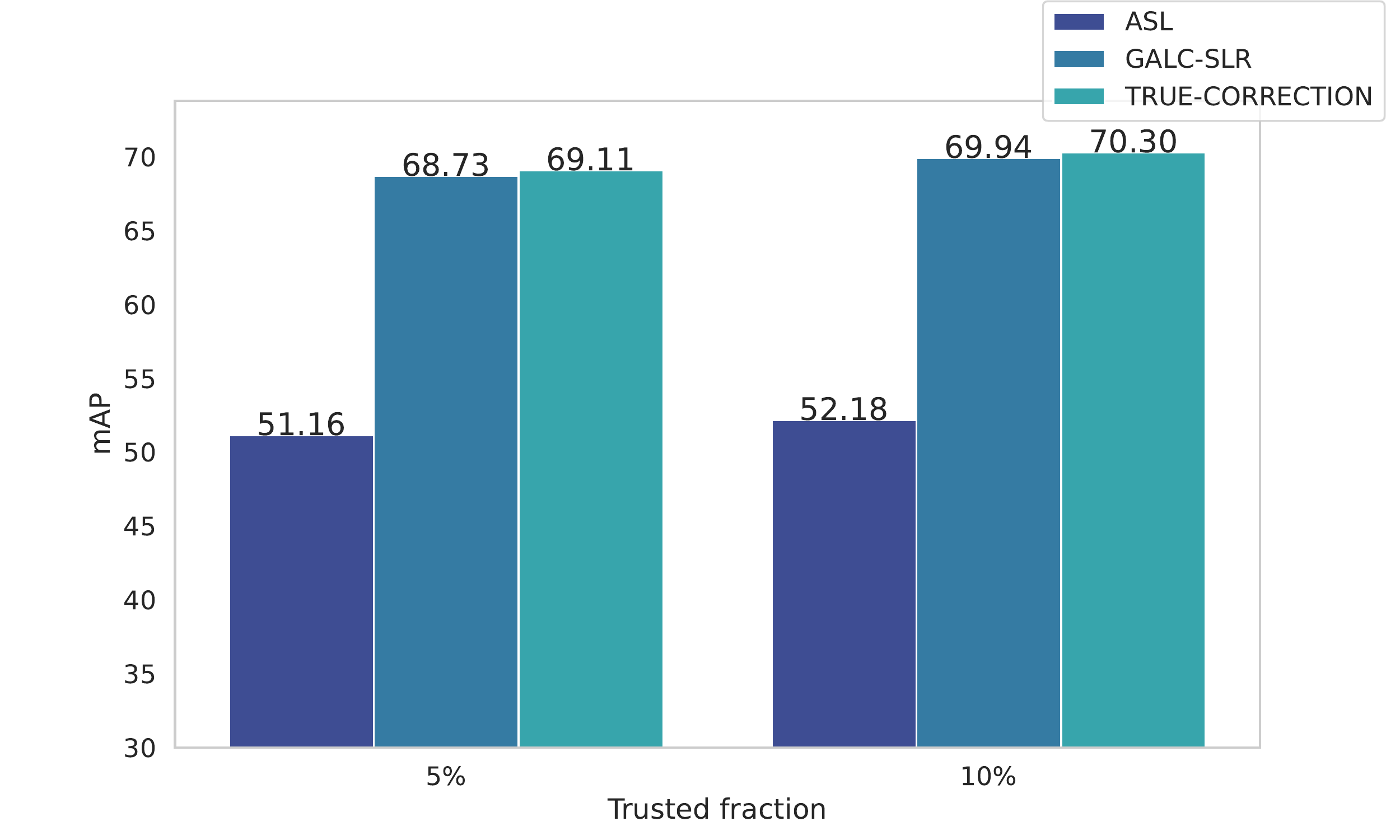}}
	\hfill
	\subfigure[Impact of single-label data]{
	    \label{subfig:ablation_study_limit_single}
	    \includegraphics[width=0.45\textwidth]{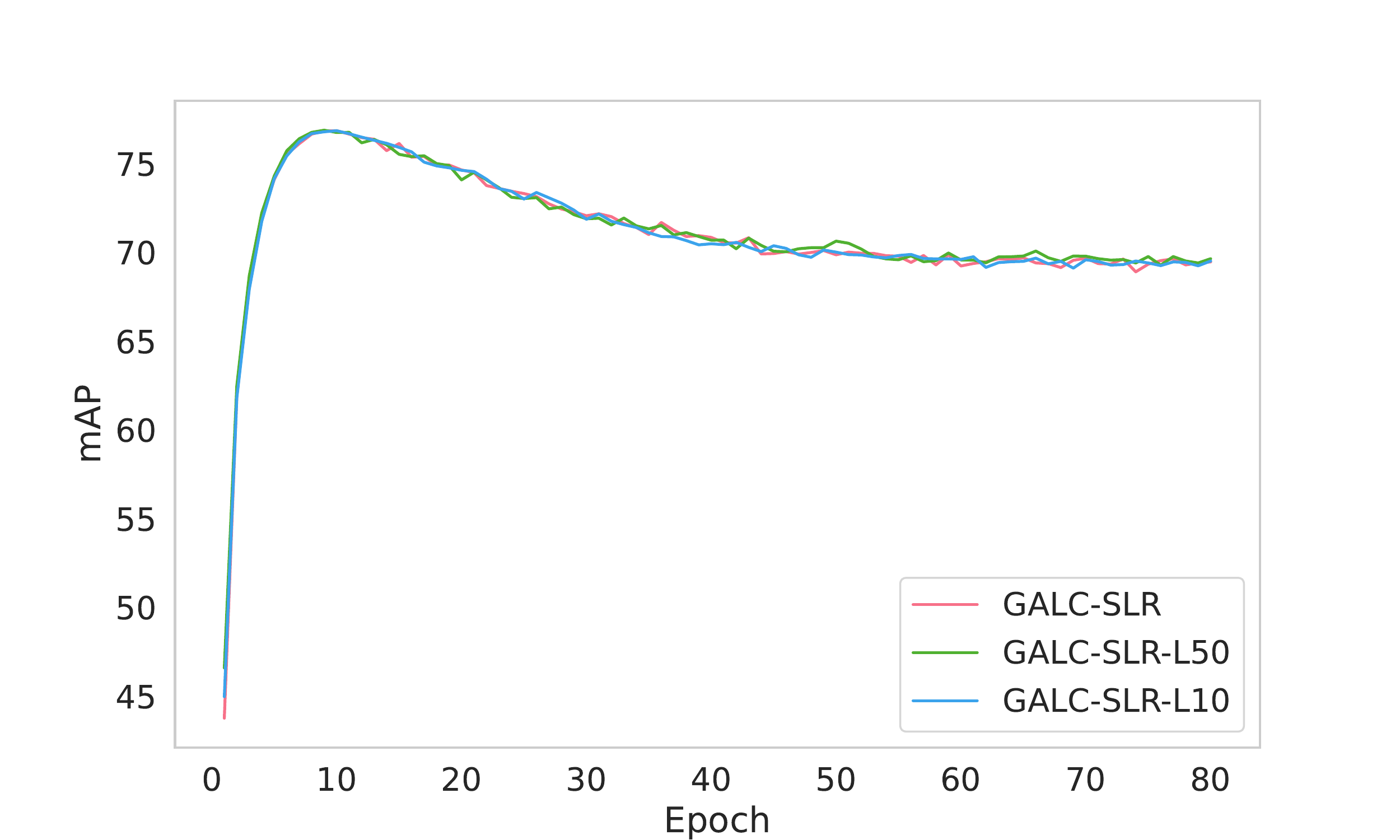}}
	\hfill
}
\caption{Ablation study of \system on MS-COCO}
\label{fig:mll_corruption_matrix}
\end{figure}

\textbf{Experiment I:} Fig.~\ref{fig:corruption_matrix} shows visually the difference between the true and our estimated noise corruption matrix. To assess also quantitatively how good our estimation method works, we train classifier $g$ with the true corruption matrix. Fig.~\ref{subfig:ablation_study_combined} compares the achieved mAP results under 40\% noise. 
$g$ trained with the true corruption matrix represents  the  upper performance bound achievable by noise corruption matrix estimators. Fig.~\ref{subfig:ablation_study_combined} shows the results across the bars of different colors.
Since \system uses trusted data to estimate the noise corruption matrix and train the robust classifier $g$, we expect the size of $\mathcal{G}$ to have an impact on the estimation accuracy and consequently on model performance. We investigate this effect by repeating the previous experiments with halve the fraction of trusted data, i.e. 5\%. This corresponds to 3,263 clean samples and 62,005 samples injected with noisy labels. Fig.~\ref{subfig:ablation_study_combined} shows these results via the two different bar plot groups.

With estimated noise corruption matrix we reach 68.73\% and 69.94\% mAP and 69.11\% and 70.30\% using the true noise correction matrix under 5\% and 10\% of trusted data, respectively. The difference between \system and the upper bound (True-Correction) is below 0.5\% points in both cases. This shows that our noise estimation is able to capture almost perfectly the impact of the noise corruption matrix, and that it works even with reduced amount of trusted data.  



\textbf{Experiment II:} To assess the impact of trusted single-label images on the estimation of the corruption matrix, we conduct two extra experiments with varying images per class.
In addition to the previous case using all single-label images, we limit the number of single-label images per class to 50 and 10, referred to as \system-L50 and \system-L10, respectively. This results in a total of  2824 for \system-L50 and 721 for \system-L10 single-label images used.

Fig.~\ref{subfig:ablation_study_limit_single} shows the impact on the mAP over training epochs. One can observe that limiting the number of single-label images has only a  minor impact on the performance of \system. Hence our proposed method is not only robust to wrong labels in multi label learning but it also can estimate an accurate noise corruption matrix by using only a small proportion of trusted single-label data. In other words, \system has a limited dependency on the amount of clean single-label data.

\section{Conclusion}
\label{sec:conclusion}
In this paper, we show the impact of wrong label information on multi-label classification. Motivated by this, we propose the Gold Asymmetric Loss Correction with Single-Label Regulators, a multi-label method that is robust against label noise. This method assumes access to a small set of clean multi-label examples as well as to a small set of clean single-label samples. \system uses this additional information in order to accurately model the label noise distribution in a multi-label setting. Through a novel regularization technique that rebalances predictions towards a targeted label, \system estimates a noise corruption matrix close to the true matrix. We evaluate \system on a real-world dataset under label noise, at multiple corruption levels from low to heavy noise. Results show that \system is a powerful method that significantly improves robustness against label noise in multi-label classification.

\bibliographystyle{unsrtnat}
\bibliography{egbib}  






\end{document}